\documentclass[sigconf]{acmart}
\usepackage{times}
\usepackage{moreverb}
\usepackage{graphicx}
\usepackage{mathrsfs}
\usepackage{bm}
\usepackage{url}
\usepackage{amsmath}
\usepackage{amssymb}

\usepackage{xspace}

\usepackage{epstopdf}
\usepackage{bbm}
\usepackage{dsfont}

\usepackage{tcolorbox}
\tcbuselibrary{most}

\usepackage{listings}

\usepackage{algorithm}
\usepackage{algorithmic}
\usepackage{balance}
\usepackage{amssymb}

\usepackage{amsmath}

\usepackage{multirow}
\usepackage{graphicx}
\usepackage{graphics}
\usepackage{lscape}
\usepackage{bm}
\usepackage{setspace}
\usepackage{verbatim}
\usepackage{todonotes}
\usepackage{url}
\usepackage{colortbl}
\usepackage{enumitem}

\newcommand{\tool}[1]{\textsc{#1}\xspace}

\newcommand{\sgd}{\tool{TestSGD }}
\newcommand{\sgde}{\tool{TestSGD}}

\AtBeginDocument{%
  \providecommand\BibTeX{{%
    \normalfont B\kern-0.5em{\scshape i\kern-0.25em b}\kern-0.8em\TeX}}}

\begin{document}
\title{
\sgde: Interpretable Testing of Neural Networks Against \\Subtle Group Discrimination
}
%


\author{Mengdi Zhang}
\email{mdzhang.2019@phdcs.smu.edu.sg}
\affiliation{%
  \institution{Singapore Management University}
  \country{Singapore}
}

\author{Jun Sun}
\email{junsun@smu.edu.sg}
\affiliation{%
  \institution{Singapore Management University}
  \country{Singapore}
}

\author{Jingyi Wang}
\email{wangjyee@zju.edu.cn}
\affiliation{%
  \institution{Zhejiang University}
  \country{China}
}

\author{Bing Sun}
\email{bing.sun.2020@phdcs.smu.edu.sg}
\affiliation{%
  \institution{Singapore Management University}
  \country{Singapore}
}



%
\begin{abstract}
Discrimination has been shown in many machine learning applications, which calls for sufficient fairness testing before their deployment in ethic-relevant domains such as face recognition, medical diagnosis and criminal sentence. Existing fairness testing approaches are mostly designed for identifying \emph{individual discrimination}, i.e., discrimination against individuals. Yet, as another widely concerning type of discrimination, testing against \emph{group discrimination, mostly hidden}, is much less studied. To address the gap, in this work, we propose \sgde, an interpretable testing approach which systematically identifies and measures hidden (which we call \emph{`subtle') group discrimination} of a neural network characterized by \emph{conditions over combinations of the sensitive features}. Specifically, given a neural network, \sgd first automatically generates an interpretable rule set which categorizes the input space into two groups exposing the model's group discrimination. Alongside, \sgd also provides an estimated group fairness score based on sampling the input space to measure the degree of the identified subtle group discrimination, which is guaranteed to be accurate up to an error bound. We evaluate \sgd on multiple neural network models trained on popular datasets including both structured data and text data. The experiment results show that \sgd is effective and efficient in identifying and measuring such subtle group discrimination that has never been revealed before. Furthermore, we show that the testing results of \sgd can guide generation of new samples to mitigate such discrimination through retraining with negligible accuracy drop.  
\end{abstract}

\maketitle

\section{Introduction}\label{sec:intro}

Machine learning models, especially neural networks, are becoming 
ubiquitous in various real-life applications. For example, they are used in medical diagnosis~\cite{kononenko2001machine}, self-driving cars~\cite{bojarski2016end} and criminal sentencing~\cite{compas2016data}. 
Meanwhile, more and more attention has been paid to the fairness issues of these machine learning models~\cite{verma2018fairness,angell2018themis,ghosh2020justicia,zhang2020white,tse2021adf,ruoss2020learning,salimi2019interventional,buolamwini2018gender,galhotra2017fairness} as discrimination has been discovered in many applications~\cite{gianfrancesco2018potential,ruoss2020learning,udeshi2018automated,zafar2017fairness}. For instance, machine learning models were used to predict recidivism risk for suspected criminals by computing the likelihood of committing a future crime~\cite{compas2016data}. Analysis results show that the prediction model was more likely to mislabel black defendants as high recidivism risk and mislabel white defendants as low risk. To minimize such ethical risks, it is crucial to systematically test the fairness of machine learning models, especially neural networks where such issues are typically `hidden' due to the lack of interpretability~\cite{pei2017deepxplore, tian2018deeptest}.

Recently, multiple efforts have been made in the testing community to first search for (and then guide mitigating) discrimination of machine learning models spanning from traditional ones to neural networks \cite{zhang2020white,tse2021adf,udeshi2018automated,zafar2017fairness,galhotra2017fairness}. For instance, state-of-the-art fairness testing work utilizes gradient information of the input sample to accelerate search/generation of discriminative samples~\cite{zhang2020white,tse2021adf,sg}. Despite being effective, existing research has mostly focused on \emph{individual discrimination}, i.e., identifying or generating individual discriminatory instances of a machine learning model~\cite{zhang2020white,tse2021adf,udeshi2018automated,zafar2017fairness,galhotra2017fairness}.
\emph{Group discrimination}, which characterizes a model's discrimination against a certain group (whose sensitive features\footnote{we use ``feature''/``attribute'' interchangeably} satisfy certain conditions), is another concerning type of discrimination, which has been widely studied~\cite{galhotra2017fairness, zafar2017fairness, tramer2017fairtest, kleinberg2016inherent}. However, testing against group discrimination has been much less studied so far. Compared to testing of individual discrimination, testing a machine learning model against group discrimination imposes new challenges. First, it is highly non-trivial to effectively enumerate all combinations of sensitive features (especially when the sensitive features have multiple or even continuous values). Second, \emph{group discrimination can be hidden, i.e., there might be `subtle' group discrimination against
those groups whose sensitive features satisfy certain unknown conditions, e.g., male-white of certain age group}. While a prior work~\cite{kearns2018preventing} similarly addresses discrimination against subgroups defined over conjunctions of protected features in the learning phase, we propose an automatic testing approach to systematically identify such subgroups using interpretable rules and measure such discrimination before model deployment.  


\begin{figure*}[t] \small
\centering 
\includegraphics[width=0.75\linewidth]{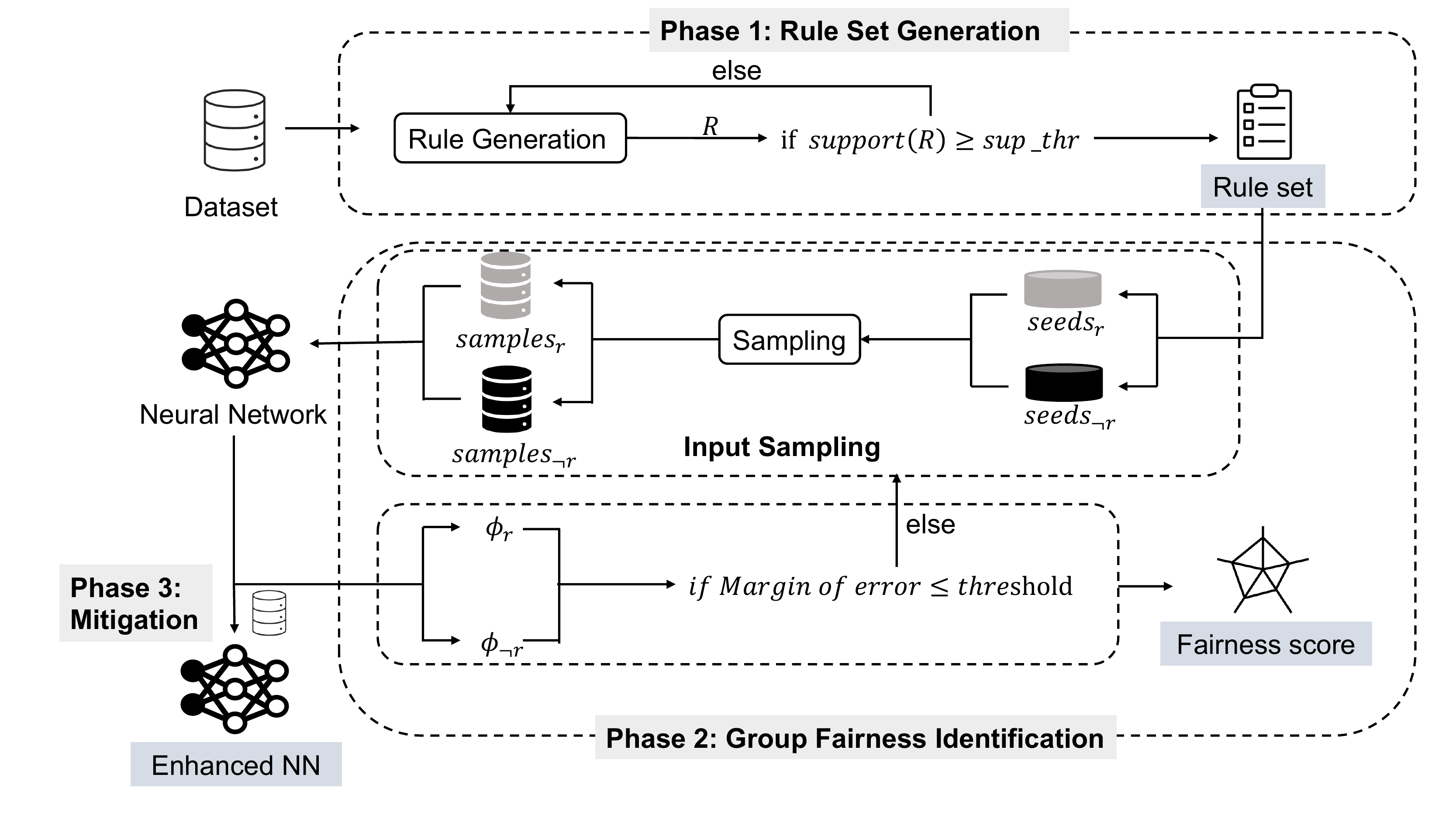}
\caption{An Overview of \sgde.} 
\label{fig:tool} 
\end{figure*}

Specifically, in this work, we develop an effective method to systematically \emph{test} a given machine learning model against such hidden \emph{s}ubtle \emph{g}roup \emph{d}iscrimination, namely \sgde. An overview of \sgd is shown in Figure~\ref{fig:tool}, which consists of three main phases: 1) candidate rule set generation, 2) group fairness identification, and 3) discrimination mitigation. In the first phase, \sgd will automatically generate a candidate set of rules concerning multiple sensitive features. Note that we only consider frequent rule set with sufficient support (which characterize a sufficiently large group). In the second phase, the rule set \emph{R} effectively partitions the samples into two groups, i.e., $samples_r$ which satisfies the rules and $samples_{\neg r}$ which does not. The key intuition behind is to develop effective criteria to automatically mine interpretable rules which are practical and relevant in the real-world applications. Then we measure if the model suffers from group discrimination (against the groups partitioned by the rule set) by measuring the group fairness score. 
Note that, solely relying on the training samples might not be enough to accurately measure such a score. We thus propose to apply a standard data augmentation method, i.e., imposing minor perturbation on the available seed samples to generate new samples, and obtain an accurate estimation of the group fairness score (with bounded errors). The testing results of the first two phases are thus the identified subtle group discrimination (characterized by the rules) and their corresponding group fairness score (with bounded errors). For example, we test the model trained on the \textbf{Crime}~\cite{crime2009dataset} dataset which predicts whether the violent crimes per population in a specific community is high. The interpretable rule set found by \sgd shows that it discriminates against communities in which the percentage of Caucasian population is lower than 80\% and the percentage of females who are divorced is higher than 40\%, with a 60.7\% group fairness score, i.e., it is 60.7\% more likely to predict high crime rate for such a community. In the last phase (optional depending whether the identified discrimination is considered to be harmful), \sgd leverages the testing results to mitigate the identified subtle group discrimination. That is, to improve group fairness, we generate new samples according to the condition under which discrimination exists and retrain the original model.      

\sgd is implemented as an open-source software~\cite{website}. We evaluate our \sgd on 8 models trained on widely adopted datasets including both structured data and text data. The experimental results show \sgd is effective in identifying and measuring subtle group discrimination. The results also show that \emph{subtle group discrimination does exist in all of these 8 models and sometimes to a surprising level which has never been revealed before}. For instance, the model trained on the \textbf{COMPAS}~\cite{compas2016data} dataset is much less likely to predict Hispanic males older than 40 years old as criminals with high recidivism risk. 
Furthermore, our experiments show that the testing-guided discrimination mitigation is useful.
That is, we can mitigate identified subtle group discrimination for all models without sacrificing the accuracy. 

In a nutshell, we summarize our main contributions as follows.
\begin{itemize} [leftmargin=*]
    \item We propose a method to automatically generate an interpretable rule set to identify 
    subtle group discrimination in neural networks, applicable for both structured and text data;
    \item We develop a theoretical bound for accurately sampling and estimating the group fairness score against two groups.  
    \item We show that we can generate samples systematically based on the interpretable rule set to mitigate subtle group discrimination.
\end{itemize}

The remainder of this paper is structured as follows. Section~\ref{sec:background} provides the background on input types and fairness definitions. Section~\ref{sec:prob_definition} defines our problem.
We present the proposed \sgd framework in Section \ref{sec:methodology}, which is evaluated in Sections \ref{sec:evaluation}. Lastly, we review related work in Section~\ref{sec:related work} and conclude in Section~\ref{sec:conclusion}.

\section{Background} \label{sec:background}
Our goal is to develop a black-box method to identify subtle group discrimination in a user-provided neural network model. Our method supports neural networks trained on two different kinds of data, i.e., structure data and text data. Our method does not require the inner details of the neural network. That is, the neural network is viewed as a function $M: {R}^{p} \to {R}^{q}$ which maps an input $x\in {R}^{p}$  to an output $y \in {R}^{q}$. Furthermore, we focus on deep feed-forward neural networks and recurrent neural networks. 

\subsection{Input Type}
First of all, we define two different data, i.e., structure data, text data, and their corresponding sensitive features which are used to evaluate the discrimination of the neural networks. 

A sample of structured dataset is composed of a set of features, i.e., a feature vector. A feature can be categorical (i.e., with a fixed set of values) or continuous (i.e., with a certain value range). We define the structure data and the corresponding sensitive features as follows. 
\begin{definition}[Structured Data] \label{def:structured data}
A structured data $x$ contains $N$ features $\{x_1, x_2, \cdots, x_N\}$, where $\forall x_i, x_i \in L_i$, where $L_i$ is a set of feature values. We write $S=\{s_1, s_2, \cdots, s_n\}$ to denote the set of sensitive features in $x$, where $n<N$. 
\end{definition}

The text data is composed of a set of tokens. We define the sensitive feature of text data based on the presence of sensitive terms. Note that there could be different categories of sensitive terms, e.g., terms referring to race, religion, or ethnicity. We define the text data and the corresponding sensitive features as follows. 

\begin{definition}[Text Data] \label{def:text data}
A text data $x$ contains a sequence of tokens $\{x_1, x_2, \cdots, x_N\}$. We write $S=\{s_1, s_2, \cdots, s_n\}$ to denote a set of categories of sensitive terms, where $n<N$ and $T$ to denote a set of sensitive terms $\{t_1, t_2, \cdots, t_k\}$, where $t_j \in s_i$ for some $i$, for all $j \in [1, k]$ and $t_j \in x$ .
\end{definition}


\subsection{Fairness Definitions} 
To define our problem, we define fairness and the concept of group fairness score. There are multiple definitions of fairness~\cite{dwork2012fairness,joseph2016fairness,calders2010three,galhotra2017fairness,kleinberg2016inherent,zafar2017fairness}. Here, we briefly review two well-studied fairness definitions, i.e., individual fairness and group fairness.

\vspace{1mm}
\noindent\textbf{Individual fairness} focuses on specific pairs of individuals. Intuitively, individual discrimination (unfairness) occurs when two individuals that differ by only certain sensitive feature (such as gender or race) are treated differently, i.e., with a different predicted label. This notion is widely used to search discriminatory instances which differ only in those sensitive characteristics~\cite{zhang2020white,tse2021adf,udeshi2018automated}. There are also plenty of works on learning models which are more likly to avoid individual discrimination~\cite{ruoss2020learning}. 

\vspace{1mm}
\noindent\textbf{Group fairness}, also known as statistical fairness, focuses on sensitive groups such as ethnic minority and the parity across different groups based on some statistical measurements~\cite{galhotra2017fairness, kleinberg2016inherent, biswas2020machine, zafar2017fairness, tramer2017fairtest}. 
A classifier satisfies this metric if the samples in the sensitive group have a positive classification probability or true positive probability that is similar with or equal to that of the insensitive group. 

In this work, we focus on group fairness for its relevance in many neural network applications. In the following, we provide a formal definition of group fairness based on positive classification rate measurement. 

\begin{definition} \label{def:group fair}
\label{def:fairness}
Let $M$ be a neural network model; $l$ be a (favorable) prediction; and $\xi$ be a positive constant. Let $G$ be a group identified by certain condition $\phi$ on sensitive features $S$. $G$ can be defined as a set of samples $\{x|x \vDash \phi\}$, where $x \vDash \phi$ means x satisfies condition $\phi$. 
We say $M$ satisfies group fairness, with respect to $\xi$ and $G$, if and only if 
\begin{equation}
\begin{aligned}
| \: P(M(x)=l \: | \: x \in G)-P(M(x)=l \: | \: x \not \in G) \: | \: \leq \xi
\end{aligned}
\end{equation}
\end{definition}

Note that, in some cases, the model may be fair overall but unfair under some specific `subtle' conditions. For example, the model is fair considering gender attribute if it approves half of the loans from female or male applicants. However, when we consider both gender and race, the model may show discrimination. For example, it approves loans for far less a percentage of Hispanic female individual, compared to the remaining group. In this setting, we say that the model discriminates against Hispanic females (if we show that the testing results have sufficient statistical confidence). 


\section{Problem Definition} \label{sec:prob_definition}
Our problem is to develop a systematic method for identifying subtle group discrimination. That is, given a neural network model $M$ (as well as a constant threshold $\xi$, we aim to generate a condition $\phi$ such that $M$ is unfair with respect to the group identified by $\phi$. The condition $\phi$ must satisfy the following conditions. (1) It must be constituted by variables representing sensitive features. (2) It must be human-interpretable, so that our analysis result can be presented for human decision making. (3) It must identify a group of non-trivial size. 
In addition, our method must support both structured data as well as text data. Furthermore, we would like our method to generate results with certain correctness guarantee, e.g., the chance of reporting non-existing discrimination is low. 

Inspired by rule-based models, which are widely used to learn interpretable models~\cite{lakkaraju2016interpretable,rivest1987learning}, we generate $\phi$ in the form of rules (a.k.a.~constraints) which are understandable by human beings and also concrete enough to show model prediction differences between different groups. The rules are constituted by the input features, without relying on any latent variables or representations. We define $\phi$ to be the conjunction of one or more rules, each of which is constituted by only one sensitive feature. Furthermore, to limit the search space as well as to make sure the generated rules are interpretable, we limit each rule on continuous features to be of the form of a linear inequality, e.g., $age \geq 30$ is a possible rule but $age~is~multiples~of~7$ is not.

In order to make sure that the discrimination that we discover is highly likely in the actual system, we propose a sampling based approach to estimate the probability of predicting certain label within a given group. Such a method allows us to generate an estimation with certain level of statistical confidence, i.e., with a bound on the error. Note that it is not straightforward to adopt existing techniques such as hypothesis testing~\cite{wald1945sequential,wald1948optimum}. This is because the group fairness score is the difference between two estimations (i.e., one for the individual in the group and the other for those not in the group). We solve this problem by establishing a conservative error bound on the difference based on the error bounds for the two estimations.  

\section{Methodology} \label{sec:methodology}
In this section, we describe the details of our approach. There are two main steps, i.e., learning a rule set and identifying group discrimination based on the learned rule set. The inputs for our method include a machine learning model $M$, its training set $D$, and a set of sensitive features $S$. The output is the subtle group discrimination represented as a rule set characterizing the discriminated group and the corresponding group fairness score. 
\subsection{Generating Frequent Rule Sets}\label{sec:rule generation}
To identify discrimination against certain group, we first need a way of characterizing a group. In this work, we characterize the groups based on a set of rules, each of which constrains one sensitive feature. In the following, we present how we generate rules for sensitive features of both structured data and text data. 

In terms of categorical features $x_i$ in structured data such as gender or race, in general, a rule can be defined as a subset of the possible feature values $L_i$. For instance, given the sensitive feature of race which has five values, i.e., Caucasian, Black, Hispanic, Asian and other-race, a rule can be any set containing one to four of these five values.
In total, we have 30 rules. 
For continuous features $x_i$ in structure data such as age or percentage, there may be too many possible values to enumerate, i.e., the domain of $L_i$ is too large. Thus we apply techniques such as binning to turn continuous features into categorical features. Here, we divide the original value range into $K$ intervals with equal width. Then we consider each interval as a single value and consider a set containing adjacent values as a rule. We set $K$ as 10 in our experiments. For example, we divide age attribute ranging from 0 to 100 into 10 equal intervals.

For textual dataset, defining rules is not that straightforward. In this work, we define the rules based on the presence of sensitive terms $T$ (refer to Definition~\ref{def:text data}). For each sensitive category $s$, we define a rule which intuitively means that the text sample contains a term $t$, where $t \in s$. In this work, we use a set of 48 terms created in~\cite{dixon2018measuring} as the sensitive terms which can be classified into 4 categories, i.e., gender, race, religion and age. The sensitive terms are shown in Table~\ref{tab:terms}. For example, when we consider the \emph{gender} feature for text dataset, there are 14 sensitive terms and thus 14 rules are generated. 

\begin{table}[t] \small
\centering
\caption{Identity Sensitive Terms}
\label{tab:terms}
\begin{tabular}{|c|l|}
\hline
\textbf{Sensitive Features}      & \multicolumn{1}{c|}{\textbf{Identity Terms}}                                                   \\ \hline
\multirow{3}{*}{gender}   & lesbian, gay, bisexual, transgender, trans,  \\
                          & queer, lgbt, lgbtq, homosexual, straight, \\
                          & heterosexual, male, female, nonbinary                           \\ \hline
\multirow{4}{*}{race}     & african, african american, black, white, \\
                          & european, hispanic, latino, latina, latinx, \\
                          & mexican, canadian, american, asian, indian,   \\
                          & middle eastern, chinese, japanese \\ \hline
\multirow{2}{*}{religion} & christian, muslim, jewish, buddhist, catholic, \\
                          & protestant, sikh, taoist, atheist                                                 \\ \hline
\multirow{2}{*}{age}      & old, older, young, younger, teenage,    \\
                          & millennial, middle aged, elderly                                                                               \\ \hline
\end{tabular}
\end{table}


Once a set of rules are identified, we then characterize a group based on a rule set. Each element of a rule set is a rule constraining one sensitive feature. Intuitively, a rule set partitions the input space of $M$ into two disjoint groups, i.e., those who satisfy all the rules in the rule set and the rest. If these two groups have a significant different probability of being predicted favorably by the model $M$, we successfully identify a subtle discrimination.   

Note that a naive approach is to enumerate all possible rules based on one sensitive feature and combine them arbitrarily. Such an approach is both infeasible and undesirable. First, there can be enormous combinations of the rules. Second, not all rule sets are interesting. For instance, a rule set may be $\{age \geq 100, gender=Male\}$. A discrimination found against the group identified by this rule set is likely to be due to the limited data. Furthermore, the discrimination is perhaps not as concerning as discrimination against groups that represent a sizable population.  

We thus only consider frequent rule sets among all possible combinations of rules. A frequent rule set is a set of rules that are satisfied by a group with a size more than certain threshold. Formally, given a rule set $R$, the \textit{support} for $R$ is the frequency of the number of samples that satisfy all rules in rule set $R$.
Given a \textit{support threshold} $\theta$ (i.e., a percentage), we say that $R$ is \emph{frequent} if its \textit{support} is no less than $\theta$. In the following, we present how to identify a set of frequent rule sets for structured and text data.  

For each sensitive feature $s$, let $R^s$ be the set of rules concerning $s$. A rule set $R$ is composed of rules for each sensitive feature, i.e., $R = \{r^{s_1}, r^{s_2}, ..., r^{s_n}\}$ where $r^{s_i} \in \{R^{s_i}\cup\varnothing\}$ and $R\neq\varnothing $. $R$ is frequent if and only if $support(R) \geq \theta$ where $support(R)$ is defined as follows. 
\begin{equation}
\label{equ:support}
\mathrm{support}(R)=\frac{\#\{d \in D | \forall r \in R.~d \vDash r\}}{\#D}  
\end{equation}
where $\#X$ of a set $X$ is the number of elements in $X$; and $d \vDash r$ means that $d$ satisfies $r$. \\


\noindent \emph{Example} Consider the structured dataset \textbf{Census Income}~\cite{census1996dataset}. It has three sensitive features, i.e., gender, race, and age. Each feature has a set of values. The following constitutes a rule set
$$\{gender = Male, race = White, 40 \leq  age < 60\}$$  \hfill $\qed$ \\
\noindent Rule sets for text data are defined differently. Recall that each rule is a proposition on whether the text contains certain sensitive term. Formally, given the set of the categories of sensitive terms $S$, a rule set $R$ is then a set of sensitive terms $\{r_1, r_2, \cdots, r_m\}$, where $r_k \in s_i$ for some $i$, for all $k \in [1,m]$ and $m\leq n$. The support of $R$ is defined as follows.
\begin{equation}
\label{equ:support}
\mathrm{support}(R)=\frac{\#\{d \in D | \forall r \in R.~contains(d,s_r)\}}{\#D}  
\end{equation}
where $s_r$ is the sensitive category referring to $r$ and $contains(d,s_{r})$ is a proposition which is true if and only if $d$ contains at least one term in the category $s_r$. \\



\noindent  \emph{Example} Consider the text dataset \textbf{Wikipedia Talk Pages}~\cite{wulczyn2017ex}. We have two categories of sensitive terms, e.g., gender and race. 
For each category, we have a set of corresponding sensitive terms as shown in Table~\ref{tab:terms}. 
The following constitutes a rule set
$$\{``bisexual", ``Caucasian"\}$$ \hfill $\qed$ \\
\noindent Algorithm~\ref{alg:ruleset} shows the exact steps in generating all possible rule sets. At line~1, we first initialize a dictionary $single\_rules$ and an empty set $rules\_sets$. During the loop from line 2 to 5, we generate all possible 1-feature rules for each sensitive feature as discussed above. At line 6, we generate a set of all 1-feature rules. Then, we generate all possible rule sets at line 7. Lastly, at line 8, we only keep those rule sets that have a support value no less than $\theta$. 

\begin{algorithm}[t]
\caption{$FrequentRuleSets(D, S, sup\_thr)$ where $D$ is the training set, $S$ is the sensitive attributes, $\theta$ is the support threshold}
\label{alg:ruleset}
\begin{algorithmic}[1]
\STATE $single\_rules \gets$ $\{\}$, $rule\_sets \gets \varnothing$
\FOR{each $s$ in $S$}
\STATE $rules \gets 
\{r_{1}, r_{2}, ...\}$
\STATE $single\_rule[s]=rules$
\ENDFOR
\STATE $all\_single\_rules \gets \{single\_rule[s] \cup \varnothing\}$ for all $s \in S$
\STATE $rule\_sets \gets combinations(all\_single\_rules)$
\STATE $all\_rule\_sets \gets \{R\ for\ R\ in\ rule\_sets\ if\ support(D, R) \geq \theta\}$
\RETURN $all\_rule\_sets$
\end{algorithmic} 
\end{algorithm}
 
In general, given a dataset with $K$ sensitive features, and at most $N$ rules for each sensitive feature, the number of rule sets is $N^{K}$ in the worse case. For example, we have 2 gender-related single rules, 5 race-related 1-feature rules and 10 age-related 1-feature rules, there are 17 rule sets when considering one sensitive attribute, 80 rule sets when considering two sensitive attributes and 100 rule sets when considering all sensitive attributes. So in total, there are 197 possible rule sets. 

\subsection{Identifying Group Fairness}
For each group identified by a rule set, we then measure the discrimination against the group. That is, we aim to compute the probability of predicting certain label by $M$ on those samples in the group, and that on those samples not in the group, and measure the difference.
The score is the group fairness score, which varies from 0 (i.e., no difference) to 1 (i.e., completely different). Formally, 

\begin{definition}[Group fairness score] \label{def:score}
Let $R$ be a rule set. Let $l$ be a (favorable) label. The group fairness score with respect to $R$ and $l$ is $|prob(R, l) - prob(\neg R, l)|$, where $prob(R,l)$ is the probability of predicting $l$ given samples satisfying $R$, $\neg R$ identifies samples not satisfying $R$.

\end{definition}

We remark that this definition is similar to the CV score~\cite{calders2010three} and multivariate group discrimination score~\cite{galhotra2017fairness}. However, the former is limited to binary input types and the latter is limited to categorical input types. In comparison, our fairness score supports both structured data and text data. \\ 

\noindent \emph{Example} Take a model trained on the (structured) \textbf{Census Income} dataset as an example. The model predicts whether the income of an adult is above \$50,000 annually, i.e., ``True'' means above the threshold and ``False'' means otherwise. 
Assume the rule set is 
$$\{gender= Male, race = White, 40 \leq age < 60\}$$ 
Assume that the model predicts 28\% of individuals in this group with ``True'', and 10\% of the remaining population with ``True''. The model's group fairness score, with respect to the rule set and the prediction, is 18\%. \hfill $\qed$ \\

\noindent 
Given a rule set, measuring the group fairness score requires us to measure $prob(R,l)$ and $prob(\neg R,l)$, which is non-trivial since exhaustively enumerating all samples is infeasible due to the enormous input space. On the other hand, measuring it based on a limited number of samples may yield inaccurate results. In the following, we propose an approach to compute group fairness scores with a statistical confidence guarantee. Formally, we would like to measure the group fairness score $f$ within a margin of error $\epsilon$ under a certain confidence $\delta$, such that $prob(|f-\hat{f}| > \epsilon) < 1-\delta$, where $\hat{f}$ is the real group fairness score over all possible samples. 

Algorithm~\ref{alg:score} shows how we measure the group fairness score. We maintain two sets of samples, i.e., $samples_r$ which contains samples satisfying $R$ and $samples_{\neg r}$ which contains samples not satisfying $R$. At line 1, we set both $samples_r$ and $samples_{\neg r}$ to be empty, error margin $\epsilon$ to be infinity and the number of generated samples as 0. During the loop from line 2 to 16, we keep generating samples and calculating group fairness score until the error margin $\epsilon$ is no more than the given error threshold $error\_thr$. From line 3 to line 6, we generate new samples for $samples_r$ and $samples_{\neg r}$ respectively using a function $Sample$. We remark that the generated samples should follow the original data distribution (i.e., that of the training dataset). We present details on how we sample on structured and text dataset in the next subsection. 

At line 7, we increase $num$ by 1. After generating a sufficient number of samples, we check the error margin $\epsilon$ from line 9 to 15. We first calculate the probability of predicting $l$ at line 9 and 10 for two sets of samples. Then at line 11, we calculate the error margin $\epsilon$ on the group fairness score. We explain why it is calculated this way below. If $\epsilon$ is less than or equal to $error\_thr$, the stopping criteria is satisfied (as in line 12 and 13). Lastly, at line 17, we return the absolute difference between $\phi_{r}$ and $\phi_{\neg r}$ as the group fairness score. 

\begin{algorithm}[t]
\caption{$GroupFairnessScore(D, M, R, sample\_thr, error\_thr)$ where $D$ is the training dataset; $M$ is the machine learning model; $R$ is a rule set, $sample\_thr$ is the number of generated inputs threshold; $error\_thr$ is error margin threshold}
\label{alg:score}
\begin{algorithmic}[1]
\STATE $samples_r \gets \varnothing$, $samples_{\neg r} \gets \varnothing$, $\epsilon \gets +\infty$, $num \gets 0$
\WHILE{$\epsilon > error\_thr$}
\STATE $x \gets Sample(D, R)$
\STATE $x' \gets Sample(D, \neg R)$
\STATE $samples_r \gets samples_r \cup {x}$
\STATE $samples_{\neg r} \gets samples_{\neg r} \cup {x'}$
\STATE $num \gets num + 1$
\IF{$num > sample\_thr$}
\STATE $\phi_r \gets \#\{i \in samples_r | M(i)=l\}/ num$
\STATE $\phi_{\neg r} \gets \#\{i \in samples_{\neg r} | M(i)=l\}/num$ 
\STATE $\epsilon \gets z\times\sqrt{\frac{\phi_r(1-\phi_r)}{num}} + z\times\sqrt{\frac{\phi_{\neg r}(1-\phi_{\neg r})}{num}} $
\IF{$\epsilon \leq error\_thr$}
\STATE break
\ENDIF
\ENDIF
\ENDWHILE
\RETURN $f \gets |\phi_r-\phi_{\neg r}|$
\end{algorithmic} 
\end{algorithm}

In the above algorithm, we estimate the error margin of the group fairness score based on an estimation of $prob(R,l)$ and $prob(\neg R,l)$. The complication is that both $prob(R,l)$ and $prob(\neg R,l)$ carry certain error margin, which may magnify the error margin for the group fairness score. In the following, we prove that line 11 in the above algorithm allows us to conservatively estimate the error margin of the group fairness score.

\begin{theorem}   \label{theorem:confidence}
Assume that $\phi_r$ satisfies the following 	
\begin{equation} \label{equ:pro}
prob(|\phi_r-\hat{\phi}_r|>\epsilon_r) < 1-\delta_r
\end{equation}
where $\epsilon_r$ and $\delta_r$ are constants. Similarly, $\phi_{\neg r}$ satisfies the following.
\begin{equation} \label{equ:unp}
prob(|\phi_{\neg r}-\hat{\phi}_{\neg r}|>\epsilon_{\neg r}) < 1-\delta_{\neg r}
\end{equation}
Then the following is satisfied. 
\begin{equation}   \label{equ:fair}
prob(|f-\hat{f}|>\epsilon_{r}+\epsilon_{\neg r}) < 1-\delta_{r}\delta_{\neg r} 
\end{equation}
 \\
\textbf{Proof:} Since $prob(|\phi_r-\hat{\phi_r}|>\epsilon_r) < 1-\delta_r$ and $prob(|\phi_{\neg r}-\hat{\phi_{\neg r}}|>\epsilon_{\neg r}) < 1-\delta_{\neg r}$, we have:
\begin{equation} 
prob(|\phi_r-\hat{\phi}_r| \leq \epsilon_r) \geq \delta_r \nonumber
\end{equation}
\begin{equation} 
prob(|\phi_{\neg r}-\hat{\phi}_{\neg r}| \leq \epsilon_{\neg r}) \geq \delta_{\neg r} \nonumber
\end{equation}
Hence 
\begin{equation} 
\begin{aligned}
&prob(|(\phi_{r}-\hat{\phi_{r}})-(\phi_{\neg r}-\hat{\phi}_{\neg r})| \leq \epsilon_{r} + \epsilon_{\neg r}) \geq \\ 
&prob(|\phi_r\!-\!\hat{\phi}_r|\!\leq\!\epsilon_r) \cdot prob(|\phi_{\neg r}\!-\!\hat{\phi}_{\neg r}|\!\leq\!\epsilon_{\neg r}) \geq \delta_r\delta_{\neg r}    \nonumber
\end{aligned}
\end{equation} 
and 
\begin{equation} 
prob(|(\phi_{r}-\hat{\phi_{r}})-(\phi_{\neg r}-\hat{\phi_{\neg r}})| > \epsilon_{r} + \epsilon_{\neg r}) < 1- \delta_r\delta_{\neg r}  \nonumber
\end{equation}
\begin{equation} 
prob(|(\phi_{r}-\phi_{\neg r})-(\hat{\phi}_{r}-\hat{\phi}_{\neg r})| > \epsilon_{r} + \epsilon_{\neg r}) < 1- \delta_r\delta_{\neg r}  \nonumber
\end{equation}
According to Definition~\ref{def:score}, group fairness score $f=\phi_{r}-\phi_{\neg r}$. Thus 
\begin{equation} 
prob(|f-\hat{f}|>\epsilon_{r}+\epsilon_{\neg r}) < 1-\delta_{r}\delta_{\neg r} \nonumber 
\end{equation} \hfill $\qed$
\end{theorem}
The above theorem provides a theoretical guarantee on the statistical confidence for the group fairness score estimation. 
That is, based on the Equation~\ref{equ:fair}, the fairness level for fairness score $f$ is $\delta_{r}\delta_{\neg r}$ and the margin of error is the sum of two margin of errors as $\epsilon_{r}+\epsilon_{\neg r}$. Each $\epsilon$ is calculated by:
\begin{equation} \label{equ:error}
\epsilon=z\times\sqrt{\frac{\phi(1-\phi)}{num}}
\end{equation}
where $z$ is the value from the standard normal distribution for a certain confidence level $\delta$ (e.g., for a 95\% confidence level, $z=1.96$). So the final margin of error for fairness score $f$ is shown in line 11 of Algorithm~\ref{alg:score}. 
Based on the result, we derive the stopping criteria, as shown in line 12 and 13 of Algorithm~\ref{alg:score}.  

The above shows how we compute the group fairness score for one rule set. Given multiple rule sets, we systematically compute the fairness score for each rule set with Algorithm~\ref{alg:score}, and then rank the rule sets according to the resultant group fairness score. If the group fairness score of certain rule set is more than a given tolerable threshold, we report that discrimination is identified. \\

\noindent \emph{Example} Take a model trained on the (structured) \textbf{Census Income} dataset as an example. We fixed the confidence level to 95\% and the corresponding $z$-value is 1.96. We set the sampling threshold $sample\_thr$ as 1000 and the error of margin threshold $error\_thr$ as 0.05. We are given a rule set 
$$\{gender = Male, race = White, 40 \leq  age < 60\}$$ 
First, we sample 1000 inputs as $samples_{r}$ using $Sample$ function that represents white males who are older than 40 but younger than 60. Then we sample another 1000 inputs as $samples_{\neg r}$ using $Sample$ function that represents the rest individuals. We observe that 283 samples in $samples_{r}$ are labeled as ``True'', while only 91 samples in $samples_{\neg r}$ are labeled as ``True''. 
So $\phi_{r}$ is 28.3\% and $\phi_{\neg r}$ is 9.1\%. According to Algorithm~\ref{alg:score}, $\epsilon_{r}$ is 0.028 and $\epsilon_{\neg r}$ is 0.018. So the margin of error $\epsilon$ for fairness score is 0.046. Since $\epsilon$ is less than 0.05, we stop sampling. Finally, the group fairness score is computed as 19.2\% with 90.25\% confidence. \hfill $\qed$

\subsection{Input Sampling}\label{sec:generation}

As discussed above, Algorithm~\ref{alg:score} requires us to sample inputs with a distribution which is similar to the data distribution of the training dataset. As shown in~\cite{goodfellow2019research}, modern machine learning models mostly rely on the i.i.d. assumptions. That is, the training and test set are assumed to be generated independently from an identical distribution. It is more likely for machine learning models to predict identically distributed data correctly. 

While it is impossible to know the actual data distribution, we aim to generate additional samples from a distribution as close as possible to the distribution of the training set. For structured data, instead of generating feature vectors randomly, we generate new samples by adding tiny perturbations on original samples uniformly. The perturbation is added to one randomly selected non-sensitive attribute with randomly selected perturbation direction and the perturbation size is 1 for integer variables or 0.01 for decimal variables. Formally, given the rule set $R$, we first search a seed instance from the dataset $D$ as $seed=\{x_1, x_2, \cdots, x_N\}$, where $\forall r \in R.~seed \vDash r$. Then we randomly select a non-sensitive attribute $x_k$, where $x_k \notin S$. We perturb $x_k$ as $x_k^{\prime} = x_k + dir \cdot s\_pert$, where $dir \in [-1, 1]$ and $s\_pert$ is the perturbation size.

For text data, we generate new samples by replacing sensitive terms with a different term in the same sensitive term category. For example, when we test the machine learning model trained on the \textbf{Wikipedia Talk Pages} dataset, given a rule set $\{``gay"\}$, we need to generate additional comments containing the term ``gay''. First, we search all comments containing gender-related sensitive terms such as ``lesbian'' and ``bisexual'', as defined in Table~\ref{tab:terms}. Then we replace these terms in the original comments with the term ``gay'' to generate new comments. That is, we can generate ``I am a gay'' from an original comment ``I am a lesbian''. The reason why we use text replacement instead of text perturbation, as in the case of structured data, is that perturbing texts with synonyms (as proposed in~\cite{sato2018interpretable} for adversarial attacks) is ineffective to generate the texts in the desired group. Our text generation method also has the benefit of mitigating the influence of data imbalance which may cause unintended bias~\cite{dixon2018measuring}. Formally, given the rule set $R=\{r_1, r_2, \cdots, r_m\}$, we first search a seed instance from the dataset $D$ as $seed=\{x_1, x_2, \cdots, x_N\}$, where $\forall r \in R.~contains(seed,s_r)$, where $s_r$ is the sensitive category referring to $r$ and $contains(d,s_{r})$ is a proposition which is true if and only if $d$ contains at least one term in the category $s_r$. Then we replace the term $x_i$ to term $r_{j}$, for all $r_{j} \in R$ and $x_i \in s_{r_{j}}$.


\section{Implementation and Evaluation} \label{sec:evaluation}

We have implemented \sgd as a self-contained software toolkit based on Tensorflow~\cite{abadi2016tensorflow} with about 6K lines of Python code. 
\vspace{1mm}
\noindent \emph{Experiment Subjects} Our experiments are based on 8 models trained with the following benchmark datasets. These datasets have been widely used as evaluation subjects in multiple previous studies on fairness~\cite{zhang2020white, tse2021adf, galhotra2017fairness, dixon2018measuring, ruoss2020learning, ma2020metamorphic}. 


\begin{itemize} [leftmargin=*]
\item{\textbf{Census Income}~\cite{census1996dataset}: 
The dataset contains more than 30,000 samples and is used to predict whether the income of an adult is above \$50,000 annually. The attributes $gender$, $race$ and $age$ are sensitive attributes. }
\item{\textbf{Bank Marketing}~\cite{moro2014data}: 
The dataset contains 45,000+ samples and is used to train models for predicting whether the client would subscribe a term deposit. Its sensitive attribute is $age$.}
\item{\textbf{German Credit}~\cite{credit1994dataset}: 
This is a small dataset with 600 samples. The task is to assess an individual’s credit. The sensitive attributes are $gender$ and $age$.}
\item{\textbf{COMPAS}~\cite{compas2016data}: 
This dataset contains 7,000+ samples. The task is to predict whether the recidivism risk score for an individual is high. The sensitive attributes are $gender$, $race$ and $age$.}
\item{\textbf{Crime}~\cite{crime2009dataset}: 
This dataset contains almost 2,000 data for communities within the US. The task is to predict whether the violent crimes per population in a specific community is high. Since this dataset records population statistics, their sensitive features are shown in multiple attributes with percentage values. Here, we extract all gender/race/age related attributes to learn rule sets.}
\item{\textbf{Law School}~\cite{anthony2003analysis}: 
This dataset has more than 20,000 application records and is used to predict whether a student passes the bar exam. The attributes, $race$ and $gender$ are sensitive attributes.}
\item{\textbf{Wiki Talk Pages}~\cite{wulczyn2017ex}: 
This is a textual dataset containing more than 100,000 Wikipedia TalkPage comments. The task is to predict whether a given comment is toxic. }
\item{\textbf{IMDB}~\cite{maas-EtAl:2011:ACL-HLT2011}: 
IMDB dataset contains 50,000 movie reviews. The task is to predict whether a given sentence is a positive review.
}
\end{itemize}

For the first six structured datasets, we train a six-layer feed-forward neural network using the exact same configuration as reported in the previous studies~\cite{zhang2020white,tse2021adf}.  For the last two textual datasets, we train a convolutional neural network (CNN) combined with long short-term memory (LSTM). The details of trained models are shown in Table~\ref{tab:models}. The accuracy of the trained models is expectedly similar to what is reported in the previous studies. Table~\ref{tab:parameters} shows the value of parameters used in our experiment to run \sgde. 
All experiments are conducted on a server running Ubuntu 1804 with 1 Intel Core 3.10GHz CPU, 32GB memory and 2 NVIDIA GV102 GPU. To mitigate the effect of randomness, all the results are the average of 3 runs. 

We aim to answer multiple research questions as follows.
\vspace{1mm}

\begin{table}[t] \small
\caption{Parameters of the Experiments}
\begin{tabular}{|c|c|c|}
\hline
Parameters & Value & Discription \\ \hline
$\theta$ & 5\% & support threshold \\ \hline
sample\_thr & 1000 & sampling threshold \\ \hline
$\delta$ & 95\% & confidence level \\ \hline
error\_thr & 0.05 & error margin threshold \\ \hline
z & 1.96 & z value \\ \hline
s\_pert & 1 & perturbation size for integer variables \\ \hline
s\_pert & 0.01 & perturbation size for decimal variables \\ \hline
\end{tabular}
\label{tab:parameters}
\end{table}

\begin{table}[t]
\caption{Dataset and Models of Experiments}
\label{tab:models}
\resizebox{\linewidth}{!}{
\begin{tabular}{|c|c|c|}
\hline
\textbf{Dataset}     & \textbf{Model}                    & \textbf{Accuracy} \\ \hline
Census Income        & Six-layer Fully-connected NN       & 86.13\%  \\ \hline
Bank Marketing       & Six-layer Fully-connected NN       & 91.62\%  \\ \hline
German Credit        & Six-layer Fully-connected NN       & 100\%    \\ \hline
COMPAS               & Six-layer Fully-connected NN       & 78.99\%  \\ \hline
Crime                & Six-layer Fully-connected NN       & 92.52\%  \\ \hline
Law School           & Six-layer Fully-connected NN       & 95.19\%  \\ \hline
Wikipedia Talk Pages & CNN Long Short-term memory network & 93.89\%  \\ \hline
IMDB                 & CNN Long Short-term memory network & 86.68\%  \\ \hline
\end{tabular}}
\end{table}

\noindent \emph{RQ1: Is our method effective in identifying subtle group discrimination of a given machine learning model?} 
\begin{table*}[t]
\caption{Rule Sets and Fairness Scores for Neural Networks}
\label{tab:results}
\resizebox{\textwidth}{!}{
\begin{tabular}{|c|cc|cc|cc|}
\hline
\multirow{3}{*}{\textbf{Dataset}} & \multicolumn{2}{c|}{\textbf{top 1}} & \multicolumn{2}{c|}{\textbf{top 2}} & \multicolumn{2}{c|}{\textbf{top 3}} \\ \cline{2-7} 
 & \multicolumn{1}{c|}{\multirow{2}{*}{\textbf{Rule Set}}} & \textbf{Fairness Score} & \multicolumn{1}{c|}{\multirow{2}{*}{\textbf{Rule Set}}} & \textbf{Fairness Score} & \multicolumn{1}{c|}{\multirow{2}{*}{\textbf{Rule Set}}} & \textbf{Fairness Score} \\  
 & \multicolumn{1}{c|}{} & ($\phi_{r}, \phi_{\neg r}$) & \multicolumn{1}{c|}{} & ($\phi_{r}, \phi_{\neg r}$) & \multicolumn{1}{c|}{} & ($\phi_{r}, \phi_{\neg r}$) \\ \hline
\multirow{2}{*}{Census Income} & \multicolumn{1}{c|}{gender=male, 40$\leq$age\textless{}80,} & 20.2\% & \multicolumn{1}{c|}{gender=male, 40$\leq$age\textless{}70,} & 19.4\% & \multicolumn{1}{c|}{gender=male, 40$\leq$age\textless{}80, race=White,} & 18.4\% \\  
 & \multicolumn{1}{c|}{race=White or Asian-Pac-Islander} & (29.9\%, 9.7\%) & \multicolumn{1}{c|}{race=White or Amer-Indian-Eskimo} & (28.9\%, 9.5\%) & \multicolumn{1}{c|}{Asian-Pac-Islander or Amer-Indian-Eskimo} & (26.9\%, 8.5\%) \\ \hline
\multirow{2}{*}{Bank Marketing} & \multicolumn{1}{c|}{\multirow{2}{*}{10 $\leq$ age < 90}} & 38.2\% & \multicolumn{1}{c|}{\multirow{2}{*}{10 $\leq$ age < 70}} & 22.8\% & \multicolumn{1}{c|}{\multirow{2}{*}{10 $\leq$ age < 60}} & 20.5\% \\ 
 & \multicolumn{1}{c|}{} & (3.3\%, 41.5\%) & \multicolumn{1}{c|}{} & (26.6\%, 3.8\%) & \multicolumn{1}{c|}{} & (4.7\%, 25.2\%) \\ \hline
\multirow{2}{*}{German Credit} & \multicolumn{1}{c|}{\multirow{2}{*}{gendre = female, 60$\leq$age\textless{}70}} & 21.9\% & \multicolumn{1}{c|}{\multirow{2}{*}{gender = female, 60$\leq$age\textless{}80}} & 21.8\% & \multicolumn{1}{c|}{\multirow{2}{*}{gender = male, 40$\leq$age\textless{}80}} & 15.5\% \\  
 & \multicolumn{1}{c|}{} & (72.5\%, 50.6\%) & \multicolumn{1}{c|}{} & (70.5\%, 48.7\%) & \multicolumn{1}{c|}{} & (52.6\%, 47.1\%) \\ \hline
\multirow{2}{*}{COMPAS} & \multicolumn{1}{c|}{gender = male, age$\geq$40,} & 62.4\% & \multicolumn{1}{c|}{gender = male, 40$\leq$age\textless{}60,} & 62.3\% & \multicolumn{1}{c|}{gender = male, 50$\leq$age\textless{}60,} & 62.3\% \\
 & \multicolumn{1}{c|}{race = Hispanic or other race} & (20.7\%, 83.1\%) & \multicolumn{1}{c|}{race = Hispanic or other race} & (20.3\%, 82.6\%) & \multicolumn{1}{c|}{race = Hispanic} & (19.5\%, 81.8\%) \\ \hline
\multirow{2}{*}{Law School} & \multicolumn{1}{c|}{\multirow{2}{*}{gender = male, race = Asian or Black}} & 15.0\% & \multicolumn{1}{c|}{\multirow{2}{*}{gender = female, race = Asian or Black}} & 11.1\% & \multicolumn{1}{c|}{\multirow{2}{*}{gender = female, race = Black}} & 10.2\% \\ 
 & \multicolumn{1}{c|}{} & (84.5\%, 99.5\%) & \multicolumn{1}{c|}{} & (88.8\%, 99.9\%) & \multicolumn{1}{c|}{} & (89.7\%, 99.9\%) \\ \hline
\multirow{2}{*}{Crime} & \multicolumn{1}{c|}{\multirow{2}{*}{FemalePctDiv$\geq$0.4, racePctWhite$\leq$0.8}} & 60.7\% & \multicolumn{1}{c|}{\multirow{2}{*}{ FemalePctDiv$\geq$0.5, racePctWhite$\leq$0.8}} & 59.6\% & \multicolumn{1}{c|}{\multirow{2}{*}{ FemalePctDiv$\geq$0.5, racePctWhite$\leq$0.6}} & 59.5\% \\  
 & \multicolumn{1}{c|}{} & (83.8\%, 23.2\%) & \multicolumn{1}{c|}{} & (87.0\%, 27.4\%) & \multicolumn{1}{c|}{} & (94.6\%, 35.1\%) \\ \hline
\multirow{2}{*}{Wiki Talk Pages} & \multicolumn{1}{c|}{\multirow{2}{*}{"gay", "taoist"}} & 6.5\% & \multicolumn{1}{c|}{\multirow{2}{*}{"gay", "protestant"}} & 5.4\% & \multicolumn{1}{c|}{\multirow{2}{*}{"gay", "african american"}} & 5.1\% \\ 
 & \multicolumn{1}{c|}{} & (13.0\%, 6.5\%) & \multicolumn{1}{c|}{} & (12.9\%, 7.5\%) & \multicolumn{1}{c|}{} & (12.5\%, 7.4\%) \\ \hline
\multirow{2}{*}{IMDB} & \multicolumn{1}{c|}{\multirow{2}{*}{"european", "yong"}} & 6.6\% & \multicolumn{1}{c|}{\multirow{2}{*}{"white", "older"}} & 6.6\% & \multicolumn{1}{c|}{\multirow{2}{*}{"lgbtq"}} & 6.5\% \\ 
 & \multicolumn{1}{c|}{} & (56.0\%, 49.4\%) & \multicolumn{1}{c|}{} & (59.1\%, 52.6\%) & \multicolumn{1}{c|}{} & (7.5\%, 14.0\%) \\ \hline
\end{tabular}}
\end{table*}
To answer the question, we systematically apply our approach to the above-mentioned models and measure the results. The results are summarized in Table~\ref{tab:results}. It shows results on the six models trained on structured data and results on the two models trained on text data. These four columns show datasets, rule sets, group fairness scores and model accuracies respectively. The favorable label is ``True'', the meaning of which can be found in the above introduction on the corresponding dataset. Note that for each model, we rank the identified subtle discrimination according to the group fairness score and we report the top 3 worst discrimination only. 

We can observe that subtle discrimination does exist in these models, which were never revealed in the previous studies~\cite{zhang2020white,tse2021adf,galhotra2017fairness,dixon2018measuring,ma2020metamorphic}. 
For instance, the model trained on the \textbf{Bank Marketing} dataset predicts only 3.3\% of the clients who are older than 10 but younger than 90 would subscribe a term deposit, whilst predicting 41.5\% of clients older than 90 would subscribe a term deposit. All of the top 3 testing results all show the model discriminates against young clients. 
We remark that although this is unfair according to the definition, it may have its underlying reasons and it is still up to human experts to decide whether it is actual discrimination. 

For the models trained on the \textbf{Census Income} dataset, \textbf{German Credit} dataset and the \textbf{Law School} dataset, they show relatively mild discrimination. In contrast, the model trained on the \textbf{COMPAS} dataset shows severe discrimination, with a fairness score of 62.4\%. That is, for Hispanic or other race male individuals who are older than 40, the model is much less likely to predict the recidivism risk as high. For the remaining individuals, the model predicts 83.1\% of them have high recidivism risk. Top 2 and top 3 test results also show severe discrimination against older Hispanic or other race male individuals. Similarly, the model trained on the \textbf{Crime} dataset also shows high discrimination. 
Different from the first five structured datasets, samples in this dataset have 10 different sensitive features, each of which is a decimal ranging from 0.0 to 1.0 representing the percentage of certain population. 
As shown in the top 1 testing result, when the percentage of divorced females is above 40\% and the percentage of Caucasian is below 80\%, the model is much more likely to predict that the violent crimes per population in this community is high. All testing results on the model trained on \textbf{Crime} dataset suggest that the model discriminates against communities with high percentage of divorced females and low percentage of Caucasian.

In Table~\ref{tab:results}, the last two rows show the results on models trained on the text data. In general, we observe that the models trained on text dataset show less discrimination. The maximum fairness score for the model trained on the \textbf{Wikipedia Talk Pages} dataset is 6.5\%. That is, the model predicts 13.0\% of comments containing both ``gay'' and ``taoist'' as toxic. For other comments (i.e., those without one of these two terms or both), the model predicts only 6.5\% of them as toxic. Top 2 and top 3 testing results show that the model discriminates against comments containing both ``gay'' and ``protestant'' and comments containing both ``gay'' and ``african american'' respectively. The model trained on the \textbf{IMDB} dataset shows a similar level of discrimination. It is more likely to predict reviewers containing ``european'' and ``young'' and reviews containing ``white'' and ``older'' as positive. It also shows discrimination against reviews containing ``lgbtq''.
Our conjecture on why the level of discrimination is considerably lower on these models is that each sample in these text datasets often has many features and as a result, the influence of each term (including sensitive terms) is distributed. 

\begin{tcolorbox}[fonttitle = \bfseries]
  \textbf{Answer to RQ1:} \sgd is effective in identifying subtle group discrimination in neural networks.
\end{tcolorbox}

\noindent \emph{RQ2: Is our method efficient?} To answer this question, we measure the amount of time required to identify the subtle discrimination for each model. The total execution time and the numbers of tested rule sets are shown in Table~\ref{tab:time}. For all models, the time required to identify the subtle discrimination is less than 20 hours. Furthermore, models trained on structured dataset take considerably less time than those trained on text dataset. That is, models trained on the \textbf{Census Income}, \textbf{Bank Marketing}, \textbf{German Credit}, \textbf{COMPAS} and \textbf{Law School} take less than 16 minutes. One exception is the model trained on the \textbf{Crime} dataset that takes more than 8 hours. The main reason is that it has a large number of rule sets, due to a large number of sensitive features (i.e., 10), all of which are continuous features. In contrast, both models trained on text dataset take more than 9 hours to finish. The main reason is that generating additional samples for such dataset takes much more time in general. We remark that the sampling procedure can be easily parallelized and thus we could significantly reduce the time if it is an issue. 

\begin{table}[t] \small
\centering
\caption{Time Taken to Identify the subtle discrimination}
\label{tab:time}
\begin{tabular}{|c|c|c|}
\hline
\textbf{Dataset} & \textbf{Time (seconds)} & \textbf{\#rule set} \\ \hline
Census Income & 869.35 & 880 \\ \hline
Bank Marketing & 141.52 & 34 \\ \hline
German Credit & 104.85 & 53 \\ \hline
COMPAS & 908.5 & 1590 \\ \hline
Law School & 18.46 & 17 \\ \hline
Crime & 29150.01 & 13282 \\ \hline
Wiki Talk Pages & 34982.28 & 732 \\ \hline
IMDB & 69125.16 & 876 \\ \hline
\end{tabular}
\end{table}

Note that the support threshold $\theta$ is set to be 5\% in all the above experiments. Intuitively, it means that each rule must be relevant to 5\% of the population (although the rule set, which is a conjunction of multiple rules, may impact a smaller population). This hyper-parameter largely determines how many rule sets that we must examine and thus may have an impact on the execution time. We thus conduct additional experiments with different $\theta$ values, ranging from 1\% to 50\%, to evaluate the effect of $\theta$ on the execution time and the results. The results on two models, i.e., the model on \textbf{Law School} and the model on \textbf{COMPAS}, are detailed in Table~\ref{tab:support thr}.

\begin{table}[t] \small
\caption{Effect of Different $\theta$}
\label{tab:support thr}
\resizebox{\linewidth}{!}{
\begin{tabular}{|c|c|c|c|c|c|}
\hline
\multirow{2}{*}{\textbf{Dataset}} & \multirow{2}{*}{\textbf{$\theta$}} & \textbf{Time} & \multirow{2}{*}{\textbf{\#rule sets}} & \multirow{2}{*}{\textbf{Rule Set}} & \textbf{Fairness} \\ 
 &  & \textbf{(seconds)} &  &  & \textbf{Score} \\ \hline
\multirow{10}{*}{Law School} & \multirow{2}{*}{1\%} & \multirow{2}{*}{46.71} & \multirow{2}{*}{59} & gender=male, & \multirow{2}{*}{16.3\%} \\ 
 &  &  &  & race=Black &  \\ \cline{2-6} 
 & \multirow{2}{*}{5\%} & \multirow{2}{*}{18.46} & \multirow{2}{*}{17} & gender=male, & \multirow{2}{*}{15.0\%} \\ 
 &  &  &  & race=Asian or Black &  \\ \cline{2-6} 
 & \multirow{2}{*}{10\%} & \multirow{2}{*}{17.83} & \multirow{2}{*}{16} & gender=male, & \multirow{2}{*}{1.0\%} \\ 
 &  &  &  & race=Asian or White &  \\ \cline{2-6} 
 & \multirow{2}{*}{20\%} & \multirow{2}{*}{17.83} & \multirow{2}{*}{16} & gender=male, & \multirow{2}{*}{0.9\%} \\ 
 &  &  &  & race=Asian or White &  \\ \cline{2-6} 
 & \multirow{2}{*}{50\%} & \multirow{2}{*}{6.28} & \multirow{2}{*}{2} & gender=male, & \multirow{2}{*}{0.3\%} \\ 
 &  &  &  & race=other race &  \\ \hline
\multirow{10}{*}{COMPAS} & \multirow{2}{*}{1\%} & \multirow{2}{*}{1175.79} & \multirow{2}{*}{2063} & gender=male, age$\geq$40 & \multirow{2}{*}{62.4\%} \\ 
 &  &  &  & race=Hispanic or other race &  \\ \cline{2-6} 
 & \multirow{2}{*}{5\%} & \multirow{2}{*}{908.50} & \multirow{2}{*}{1590} & gender=male, age$\geq$40 & \multirow{2}{*}{62.4\%} \\ 
 &  &  &  & race=Hispanic or other race &  \\ \cline{2-6} 
 & \multirow{2}{*}{10\%} & \multirow{2}{*}{676.74} & \multirow{2}{*}{1180} & gender=male, age$\geq$20 & \multirow{2}{*}{43.9\%} \\ 
 &  &  &  & race=Hispanic or other race &  \\ \cline{2-6} 
 & \multirow{2}{*}{20\%} & \multirow{2}{*}{0} & \multirow{2}{*}{0} & \multirow{2}{*}{NULL} & \multirow{2}{*}{NULL} \\
 &  &  &  &  &  \\ \cline{2-6} 
 & \multirow{2}{*}{50\%} & \multirow{2}{*}{0} & \multirow{2}{*}{0} & \multirow{2}{*}{NULL} & \multirow{2}{*}{NULL} \\
 &  &  &  &  &  \\ \hline
\end{tabular}}
\end{table}

The table shows the execution time, the number of rule sets and the worst group fairness score. We can observe that, the larger a $\theta$ we set, the fewer rule sets, the less execution time and the smaller group fairness score in general. If the threshold $\theta$ is too low, e.g., 1\%, we spend a lot of time on testing a huge number of rule sets, which may not be interesting (one such example is $\{gender=Male, age \geq 100\}$). In contrast, if the threshold $\theta$ is too high, e.g., 20\% or 50\%, there may only exists few or even none rule set (as in the case of the model trained on the \textbf{COMPAS} dataset).  

We note that different $\theta$ may result in different discrimination being identified. For the model trained on \textbf{Law School}, the rule set shows that the model discriminates against black or Asian males the most when $\theta$ is 5\%. However, when we set $\theta$ to be 1\%, the model is shown to discriminate against black male individuals the most. For the model trained on the \textbf{COMPAS} dataset, the model discriminates against Hispanic or other race males who is older than 40 years old most when we set $\theta$ to be 5\%. However, when we set $\theta$ higher (i.e., 10\%), the age range is expanded to be over 20 years in the identified rule set. Such a result is expected as a large $\theta$ requires us to find discrimination against a large group. What is considered to be a reasonable value for $\theta$ is a complicated question, which should probably be answered by lawmakers.

\begin{tcolorbox}[fonttitle = \bfseries]
  \textbf{Answer to RQ2:} \sgd is reasonably efficient.
\end{tcolorbox}


\begin{table*}[t] \small
\centering
\caption{Discrimination Mitigation for Neural Networks}
\label{tab:mitigation}
 \resizebox{0.65\linewidth}{!}{
\begin{tabular}{|c|c|cc|cc|}
\hline
\multirow{3}{*}{\textbf{Dataset}} & \multirow{3}{*}{\textbf{Rule Set}} & \multicolumn{2}{c|}{\textbf{Before}} & \multicolumn{2}{c|}{\textbf{After}} \\ \cline{3-6} 
 &  & \multicolumn{1}{c|}{\multirow{2}{*}{\textbf{accuracy}}} & \textbf{Fairness Score} & \multicolumn{1}{c|}{\multirow{2}{*}{\textbf{accuracy}}} & \textbf{Fairness Score} \\   
 &  & \multicolumn{1}{c|}{} & ($\phi_{r}, \phi_{\neg r}$) & \multicolumn{1}{c|}{} & ($\phi_{r}, \phi_{\neg r}$) \\ \hline
\multirow{2}{*}{Census Income} & gender=male, 40$\leq$age<80, & \multicolumn{1}{c|}{\multirow{2}{*}{86.1\%}} & 20.2\% & \multicolumn{1}{c|}{\multirow{2}{*}{86.2\%}} & 10.1\% \\ 
 & race=White or Asian-Pac-Islander & \multicolumn{1}{c|}{} & (29.9\%, 9.7\%) & \multicolumn{1}{c|}{} & (18.9\%, 8.8\%) \\ \hline
\multirow{2}{*}{Bank Marketing} & \multirow{2}{*}{10 $\leq$ age < 90} & \multicolumn{1}{c|}{\multirow{2}{*}{91.6\%}} & 38.2\% & \multicolumn{1}{c|}{\multirow{2}{*}{90.6\%}} & 5.4\% \\ 
 &  & \multicolumn{1}{c|}{} & (3.3\%, 41.5\%) & \multicolumn{1}{c|}{} & (6.9\%, 12.3\%) \\ \hline
\multirow{2}{*}{German Credit} & \multirow{2}{*}{gender = female, 60$\leq$age<70} & \multicolumn{1}{c|}{\multirow{2}{*}{100.0\%}} & 21.9\% & \multicolumn{1}{c|}{\multirow{2}{*}{100.0\%}} & 7.3\% \\ 
 &  & \multicolumn{1}{c|}{} & (72.3\%, 50.6\%) & \multicolumn{1}{c|}{} & (45.9\%, 53.2\%) \\ \hline
\multirow{2}{*}{COMPAS} & gender = male, age$\geq$40, & \multicolumn{1}{c|}{\multirow{2}{*}{79.0\%}} & 62.4\% & \multicolumn{1}{c|}{\multirow{2}{*}{78.5\%}} & 4.2\% \\ 
 & race = Hispanic or other race & \multicolumn{1}{c|}{} & (20.7\%, 83.1\%) & \multicolumn{1}{c|}{} & (80.9\%, 85.1\%) \\ \hline
\multirow{2}{*}{Law School} & \multirow{2}{*}{gender = male, race = Black} & \multicolumn{1}{c|}{\multirow{2}{*}{95.2\%}} & 15.0\% & \multicolumn{1}{c|}{\multirow{2}{*}{95.1\%}} & 7.5\% \\
 &  & \multicolumn{1}{c|}{} & (84.5\%, 99.5\%) & \multicolumn{1}{c|}{} & (92.3\%, 99.8\%) \\ \hline
\multirow{2}{*}{Crime} & \multirow{2}{*}{FemalePctDiv$\geq$0.4, racePctWhite$\leq$0.8} & \multicolumn{1}{c|}{\multirow{2}{*}{93.9\%}} & 60.7\% & \multicolumn{1}{c|}{\multirow{2}{*}{98.1\%}} & 51.4\% \\ 
 &  & \multicolumn{1}{c|}{} & (83.8\%, 23.2\%) & \multicolumn{1}{c|}{} & (90.6\%, 39.2\%) \\ \hline
\multirow{2}{*}{Wiki Talk Pages} & \multirow{2}{*}{"gay", "taoist"} & \multicolumn{1}{c|}{\multirow{2}{*}{93.9\%}} & 6.5\% & \multicolumn{1}{c|}{\multirow{2}{*}{95.5\%}} & 0.4\% \\ 
 &  & \multicolumn{1}{c|}{} & (13.0\%, 6.5\%) & \multicolumn{1}{c|}{} & (8.4\%, 8.0\%) \\ \hline
\multirow{2}{*}{IMDB} & \multirow{2}{*}{"european", "yong"} & \multicolumn{1}{c|}{\multirow{2}{*}{86.7\%}} & 6.6\% & \multicolumn{1}{c|}{\multirow{2}{*}{84. \%}} & 3.3\% \\ 
 &  & \multicolumn{1}{c|}{} & (56.0\%, 49.4\%) & \multicolumn{1}{c|}{} & (43.7\%, 40.4\%) \\ \hline
\end{tabular}}
\end{table*}

\noindent \emph{RQ3: Can we mitigate subtle discrimination using our testing results?} To further show the usefulness of our approach, we evaluate whether we can mitigate the identified subtle discrimination using our testing results. The idea is to mitigate the discrimination by retraining. We remark that there are alternative approaches for improving fairness as well~\cite{kearns2018preventing,salimi2019interventional}. Note that we generate additional instances satisfying the rule set with the sampling approach described in Section~\ref{sec:generation}. We only select those generated instances with the opposite label. For example, the model trained on \textbf{COMPAS} is more likely to predict elderly males who are Hispanic or other race with ``False'' label. We can use the $Sample$ function to generate instances satisfying the condition that are labeled as ``True'' according to the original model. Afterward, we retrain the original model with these additional instances and testing the subtle discrimination with respect to the same rule set to see the improvement. Note that we gradually increase the number of additional instances from 50 to 10\% of original dataset size to achieve the lowest fairness score without decreasing the accuracy of the retrained model.

We only consider the top 1 worst rule sets to mitigate the discrimination. The results are shown in Table~\ref{tab:mitigation} for six models trained on additional structured data and two models retrained on additional textual data. We can observe that all models show reduced subtle discrimination and almost the same accuracy. The fairness scores for retrained models on \textbf{Census Income}, \textbf{German Credit} and \textbf{Law School} decrease by about half. For the most improvement, the model retrained on the \textbf{COMPAS} dataset shows much less subtle discrimination as the fairness score decreases by more than 10 times, i.e., from 57.7\% to 4.2\%. The fairness score of the model trained on the \textbf{Crime} dataset decreases from 60.7\% to 51.4\%. Relatively, the fairness improvement is not obvious. We believe that it is due to its many continuous sensitive features and the large number of features (i.e., each input contains more than 100 attributes). That is, it would require a lot more additional data to improve fairness. In terms of CNN models, the fairness score decreases from from 6.5\% to 0.4\% for the model retrained on \textbf{Wikipedia Talk Pages} and decreases from 6.6\% to 3.3\% for the model retrained on \textbf{IMDB}. 
\begin{tcolorbox}[fonttitle = \bfseries]
  \textbf{Answer to RQ3:} \sgde$ $ is useful in mitigating the identified subtle group discrimination through retraining.
\end{tcolorbox}

\noindent \emph{Comparison with Baselines} We identify the following two baselines from literature which can potentially identify similar group discrimination as our work.
\textbf{1) THEMIS~\cite{galhotra2017fairness}} 
calculates group discrimination scores over combinations of multiple features (subgroups) by measuring the difference between the maximum and minimum frequencies of two subgroups on randomly generated samples. 
Those subgroups 
can then be regarded as identified discrimination if the score is higher than a threshold. \textbf{2) FairFictPlay \cite{kearns2018preventing}} proposed an in-processing algorithm aiming to improve subgroup fairness. The subgroups are identified with user-provided constraints in the form of conjunctions of Boolean attributes, linear threshold functions, or bounded degree polynomial threshold functions over multiple protected features. 


In Table~\ref{tab:baseline}, we show the identified group discrimination with \sgde, Themis and FairFictPlay respectively, along with the fairness scores,
on the same models trained on structured data (similarly to  
Table~\ref{tab:models}). 
We set the timeout as 24 hours. Note that FairFictPlay uses complex linear functions on all the protected features (which are hard to interpret) to define discriminatory subgroups, thus 
we do not show the exact concrete linear functions in the table. 
We have the following observations. 1) Compared to FairFictPlay, \sgd identifies discrimination with higher scores (more discriminating) while being interpretable. Moreover, \sgd automatically identifies the discriminated subgroups without any prior knowledge. 
2) Similar to \sgde, Themis is able to identify discriminated subgroups automatically. However, Themis identifies two subgroups which are maximally different (in terms of being predicted favorably) while \sgd identifies subgroups which are predicted different from the rest. These two approaches thus produce results that are complementary to each other. Note that Themis does not support text data.   

\begin{table*}[t]
\caption{Comparisons Between \sgd, Themis and FairFictPlay. `-' means timeout.}
\label{tab:baseline}
\resizebox{\linewidth}{!}{
\begin{tabular}{|c|cc|cc|cc|}
\hline
\multirow{2}{*}{\textbf{Dataset}} & \multicolumn{2}{c|}{\textbf{\sgd}} & \multicolumn{2}{c|}{\textbf{Themis}} & \multicolumn{2}{c|}{\textbf{FairFictPlay}} \\ \cline{2-7} 
 & \multicolumn{1}{c|}{\textbf{Rule Set}} & \textbf{Fairness Score} & \multicolumn{1}{c|}{\textbf{Sensitive Attributes' values for Max/Min Proportion}} & \textbf{Fairness Score} & \multicolumn{1}{c|}{\textbf{Subgroup}} & \textbf{Fairness Score} \\ \hline
\multirow{2}{*}{Census Income} & \multicolumn{1}{c|}{Gender=male, 40$\leq$age<80,} & \multirow{2}{*}{20.20\%} & \multicolumn{1}{c|}{{[}gender=Female, 60$\leq$age\textless{}70, race=Asian-Pac\_islander{]} -} & \multirow{2}{*}{26.6\%} & \multicolumn{1}{c|}{\multirow{2}{*}{Linear Threshold Function}} & \multirow{2}{*}{13.9\%} \\
 & \multicolumn{1}{c|}{race=White or Asian-Pac\_islander} &  & \multicolumn{1}{c|}{{[}gender=Male, 10$\leq$age\textless{}20, race=White{]}} &  & \multicolumn{1}{c|}{} &  \\ \hline
\multirow{2}{*}{Bank Marketing} & \multicolumn{1}{c|}{\multirow{2}{*}{10$\leq$age\textless{}90}} & \multirow{2}{*}{38.2\%} & \multicolumn{1}{c|}{\multirow{2}{*}{{[}60$\leq$age\textless{}70{]} - {[}10$\leq$age\textless{}20{]}}} & \multirow{2}{*}{8.4\%} & \multicolumn{1}{c|}{\multirow{2}{*}{Linear Threshold Function}} & \multirow{2}{*}{7.6\%} \\
 & \multicolumn{1}{c|}{} &  & \multicolumn{1}{c|}{} &  & \multicolumn{1}{c|}{} &  \\ \hline
\multirow{2}{*}{German Credit} & \multicolumn{1}{c|}{\multirow{2}{*}{gender=feamle, 60$\leq$age\textless{}70}} & \multirow{2}{*}{21.9\%} & \multicolumn{1}{c|}{{[}gender=Female, 80$\leq$age\textless{}90{]} -} & \multirow{2}{*}{17.1\%} & \multicolumn{1}{c|}{\multirow{2}{*}{Linear Threshold Function}} & \multirow{2}{*}{7.0\%} \\
 & \multicolumn{1}{c|}{} &  & \multicolumn{1}{c|}{{[}gender=Male, 10$\leq$age\textless{}20{]}} &  & \multicolumn{1}{c|}{} &  \\ \hline
\multirow{2}{*}{COMPAS} & \multicolumn{1}{c|}{gender=male, age$\geq$40,} & \multirow{2}{*}{62.4\%} & \multicolumn{1}{c|}{{[}gender=Female, 10$\leq$age\textless{}20, race=Native American{]} -} & \multirow{2}{*}{67.3\%} & \multicolumn{1}{c|}{\multirow{2}{*}{Linear Threshold Function}} & \multirow{2}{*}{22.4\%} \\
 & \multicolumn{1}{c|}{race=Hispanic or other race} &  & \multicolumn{1}{c|}{{[}gender=Male, 60$\leq$age\textless{}70, race=other race{]}} &  & \multicolumn{1}{c|}{} &  \\ \hline
\multirow{2}{*}{Law School} & \multicolumn{1}{c|}{\multirow{2}{*}{gender=male, race=Asian or Black}} & \multirow{2}{*}{15.0\%} & \multicolumn{1}{c|}{{[}gender=Male, race=White{]} -} & \multirow{2}{*}{13.5\%} & \multicolumn{1}{c|}{\multirow{2}{*}{Linear Threshold Function}} & \multirow{2}{*}{3.7\%} \\
 & \multicolumn{1}{c|}{} &  & \multicolumn{1}{c|}{{[}gender=Female, race=Black{]}} &  & \multicolumn{1}{c|}{} &  \\ \hline
\multirow{2}{*}{Crime} & \multicolumn{1}{c|}{\multirow{2}{*}{FemalePctDiv$\geq$0.4, racePctWhite$\leq$0.8}} & \multirow{2}{*}{60.7\%} & \multicolumn{1}{c|}{\multirow{2}{*}{-}} & \multirow{2}{*}{-} & \multicolumn{1}{c|}{\multirow{2}{*}{Linear Threshold Function}} & \multirow{2}{*}{38.8\%} \\
 & \multicolumn{1}{c|}{} &  & \multicolumn{1}{c|}{} &  & \multicolumn{1}{c|}{} &  \\ \hline
\end{tabular}}
\end{table*}

\vspace{-1em}
\section{Related Work} \label{sec:related work}
Many existing works attempted to test discrimination according to different fairness definitions and measurements~\cite{dwork2012fairness, calders2010three}. In~\cite{feldman2015certifying}, Feldman \emph{et al.} provide a fairness definition which is measured according to demographic parity of model predictions. It measures how well the sensitive class can be predicted based on classification accuracy. In~\cite{hardt2016equality}, Hardt \emph{et al.} present an alternate definition of fairness based on demographic parity. It requires a decision to be independent of the sensitive attribute. In~\cite{kusner2017counterfactual}, Kusner \emph{et al.} define counterfactual discrimination which focuses on single decisions towards an individual. A prediction is counterfactual fair if it is the same in the actual group and a different demographic group. In~\cite{galhotra2017fairness}, Galhotra \emph{et al.} propose causal discrimination to measure the fraction of inputs for which model causally discriminates. This definition is similar to counterfactual fairness, but it takes instances of discrimination into account. In~\cite{kearns2018preventing}, Kearns \emph{et al.} proposed an in-processing algorithm aiming to improve the fairness of given subgroups, where subgroups are defined as conjunctions of attributes, linear threshold functions, or bounded degree polynomial threshold functions over multiple protected features.
Most existing works~\cite{galhotra2017fairness, kleinberg2016inherent, biswas2020machine} use positive classification rate as fairness measurement. 
 
Subsequently, many works focus on individual discrimination to generate individual discriminatory instances~\cite{zhang2020white,tse2021adf,agarwal2018automated, huchard2018proceedings}. They tried to generated instances which are classified differently after changing sensitive attributes. 
In~\cite{agarwal2018automated}, Agarwal \emph{et al.} present an automated testing approach to generate test inputs to find individual discrimination. In~\cite{ruoss2020learning}, Ruoss \emph{et al.} propose a fairness representation framework to generalize individual fairness to multiple notions. It learns a mapping from similar individuals to latent representations.
However, the testing on individual discrimination cannot provide a statistical measurement of fairness.  

Some other existing works attempted to test model discrimination with fairness score measurements. In~\cite{tramer2017fairtest}, Tramer \emph{et al.} propose an unwarranted associations framework to detect unfair, discriminatory or offensive user treatment in data-driven applications. It identifies discrimination according to multiple metrics including the CV score, related ratio and associations between outputs and sensitive attributes. In~\cite{kleinberg2016inherent}, Kleinberg \emph{et al.} also test multiple discrimination scores and compare different fairness metrics. In~\cite{galhotra2017fairness}, Galhotra \emph{et al.} propose a tool called THEMIS to measure software discrimination. It tests discrimination with two fairness definitions, i.e., group discrimination score and causal discrimination score. 
In~\cite{adebayo2016iterative}, Adebayo \emph{et al.} try to determine the relative significance of a model's inputs in determining the outcomes and use it to assess the discriminatory extent of the model. 

Some prior work has been done on fairness for text classification tasks as well. In~\cite{blodgett2017racial}, Blodgett \emph{et al.} discuss the impact of unfair natural language in NLP and show how statistical discrimination arises in processing applications. 
In~\cite{bolukbasi2016man}, Bolukbasi \emph{et al.} show gender bias in the world embedding and provide a methodology for modifying an embedding to remove gender bias. In~\cite{dixon2018measuring}, Dixon \emph{et al.} measure discrimination using a set of common demographic identity terms and propose a method to mitigate the unintended bias by balancing the training data. 

Compared with all the above-mentioned existing works, we provide further fairness testing. Instead of measuring the overall discrimination, our approach systematically identifies and measures subtle discrimination. That is, we not only measure statistical discrimination with a confidence guarantee but also offer interpretable rule sets to represent subtle discrimination. 

This work is remotely related to works on applying rule-based models for model explanation. In~\cite{yang2017scalable}, Yang \emph{et al.} present an algorithm for building probabilistic rule lists with logical IF-THEN structure.
In~\cite{lakkaraju2016interpretable}, Lakkaraju \emph{et al.} propose interpretable decision sets to interpret model predictions with high accuracy and high interpretation. Our work leverage such rule-based interpretable structure to present subtle discrimination in models. 

\vspace{-1em}
\section{Conclusion} \label{sec:conclusion}
In this work, we focus on testing neural network models against subtle group discrimination and propose a framework to systematically identify interpretable subtle group discrimination based on group fairness measurement with a certain confidence. 
Our extensive evaluation demonstrates that subtle group discrimination in neural networks is common to a surprising level. 
We also show that it is possible to mitigate such discrimination by utilizing our testing results to generate more data for retraining. 

%
%
%

\balance
\bibliographystyle{splncs04}
\bibliography{ref}

\end{document}